\documentclass[final]{IEEEtran}
\IEEEoverridecommandlockouts
\usepackage[T1]{fontenc}
\usepackage{graphicx}
\usepackage{hyperref}
\usepackage{color}
\usepackage{xspace}
\usepackage{amsmath,amsfonts,amssymb}
\usepackage{booktabs}

\usepackage{marvosym}
\usepackage{tabularx}
\usepackage{subfig}
\usepackage{siunitx}
\usepackage{multirow, makecell}
\usepackage{todonotes}
\usepackage{mathtools}
\usepackage{tikz}
\usepackage{comment}
\usepackage[most]{tcolorbox}
\usepackage{float}
\usepackage{svg}
\usepackage{orcidlink}

\newtcolorbox{Summary}{
    fontupper = \color{black}, 
    boxrule = 1pt,
    colback = black!2!white,
    colframe = black!70!white,
    rounded corners,
    arc = 5pt   
}

\usetikzlibrary{arrows.meta, positioning, backgrounds, fit, calc, arrows.meta}
\usepackage{pgfplots}
\usepackage{comment}
\usepgfplotslibrary{fillbetween}
\definecolor{mygray}{RGB}{96, 125, 139}
\definecolor{myred}{RGB}{211, 47, 47}
\definecolor{mypanel}{RGB}{248, 250, 252}
\definecolor{myredbg}{RGB}{252, 232, 232}
\definecolor{myblue}{RGB}{25, 118, 210}      
\definecolor{mybluebg}{RGB}{227, 242, 253}    
\definecolor{mygreen}{RGB}{46, 125, 50}        
\definecolor{mygreenbg}{RGB}{232, 245, 233}    
\definecolor{mypurple}{RGB}{106, 27, 154}      
\definecolor{mypurplebg}{RGB}{243, 229, 245}   
\definecolor{mygray}{RGB}{128, 134, 139}  

\begin{document}

\newcommand{\dbra}[1]{\ensuremath{\llbracket #1 \rrbracket}} 

\NewDocumentCommand{\opinion}{m o}{%
  \ensuremath{\omega_{#1}\IfValueT{#2}{^{#2}}}%
}
\NewDocumentCommand{\opset}{m o}{%
  \ensuremath{\boldsymbol{\omega}_{#1}\IfValueT{#2}{^{#2}}}%
}
\NewDocumentCommand{\binop}{m}{\ensuremath{\langle b_{#1},\, d_{#1},\, u_{#1},\, a_{#1}\rangle}} 

\newcommand{\belief}[1]{\ensuremath{b_{#1}}}
\newcommand{\disbelief}[1]{\ensuremath{d_{#1}}}
\newcommand{\uncert}[1]{\ensuremath{u_{#1}}}
\newcommand{\baserate}[1]{\ensuremath{a_{#1}}}

\newcommand{\vbelief}[1]{\ensuremath{\mathbf{b}_{#1}}}
\newcommand{\vbaserate}[1]{\ensuremath{\mathbf{a}_{#1}}}
\newcommand{\uncertV}[1]{\ensuremath{u_{#1}}}

\newcommand{\marginalop}[1]{\dbra{#1}}

\newcommand{\deduces}{\mathrel{\Vert}} 
\newcommand{\deducedopinion}[2]{\opinion{#1\deduces #2}}

\newcommand{\given}{\mathrel{\mid}}   
\newcommand{\conditionalopinion}[2]{\opinion{#1\given #2}}
\NewDocumentCommand{\conditionals}{s m m}{%
\ensuremath{
  \left( \omega_{#2 \mid #3},\, \omega_{#2 \mid \lnot\, \IfBooleanTF{#1}{(#3)}{#3}} \right)%
}
}

\newcommand{\proj}[1]{\ensuremath{p_{#1}}}
\newcommand{\projof}[1]{\ensuremath{\operatorname{proj}\!\left(#1\right)}}
\newcommand{\expect}[1]{\ensuremath{\mathbb{E}\!\left[#1\right]}}
\newcommand{\expecop}[1]{\ensuremath{\mathbb{E}\!\left[\opinion{#1}\right]}}

\newcommand{\projrule}[1]{\ensuremath{\belief{#1} + \baserate{#1}\,\uncert{#1}}} 
\newcommand{\normop}[1]{\ensuremath{\belief{#1} + \disbelief{#1} + \uncert{#1} = 1}}

\newcommand{\evidence}[1]{\ensuremath{\mathbf{r}_{#1}}}
\newcommand{\evidp}[1]{\ensuremath{r_{#1}^{+}}}
\newcommand{\evidm}[1]{\ensuremath{r_{#1}^{-}}}
\newcommand{\dirichlet}[1]{\ensuremath{\operatorname{Dir}\!\left(#1\right)}}
\newcommand{\BetaDist}[1]{\ensuremath{\operatorname{Beta}\!\left(#1\right)}}
\newcommand{\alphaDir}[1]{\ensuremath{\boldsymbol{\alpha}_{#1}}}

\newcommand{\cumfusion}{\ensuremath{\mathbin{\oplus}}}        
\newcommand{\afusion}{\ensuremath{\mathbin{\underline{\oplus}}}} 
\newcommand{\constraintfusion}{\ensuremath{\mathbin{\odot}}} 
\newcommand{\discount}{\ensuremath{\mathbin{\otimes}}}        
\newcommand{\deduction}{\ensuremath{\mathbin{\circledcirc}}}  
\newcommand{\abduction}{\ensuremath{\tilde{\mathbin{\circledcirc}}}} 
\newcommand{\inversion}{\ensuremath{\tilde{\phi}}}                
\newcommand{\weightedfusion}{\ensuremath{\mathbin{\hat{\oplus}}}} 
\newcommand{\ccfusion}{\ensuremath{\mathbin{\ooalign{%
  \hfil \raisebox{0.1ex}{\tiny CC}\hfil\crcr
  $\bigcirc$}}}}  
\newcommand{\unfusion}{\ensuremath{\mathbin{\ominus}}}        

\newcommand{\negate}[1]{\ensuremath{\overline{#1}}}
\newcommand{\co}[1]{\ensuremath{#1^{\complement}}}
\newcommand{\domain}[1]{\ensuremath{\mathbb{#1}}}
\newcommand{\variable}[1]{\ensuremath{\mathcal{#1}}}

\newtheorem{definition}{Definition}

\newtcolorbox{RQSummary}{
    fontupper = \color{black}, 
    boxrule = 1pt,
    colback = black!2!white,
    colframe = black!70!white,
    rounded corners,
    arc = 5pt   
}
%

%
\title{CUBICS: Situation-aware performance estimation for safety‑relevant ML components}

\author{
\IEEEauthorblockN{Benjamin Herd\orcidlink{0000-0001-6439-8845}\IEEEauthorrefmark{1}, Jessica Kelly\orcidlink{0009-0003-0508-2367}\IEEEauthorrefmark{1}, and Mario Trapp\orcidlink{0000-0002-4716-7372}\IEEEauthorrefmark{2} \IEEEauthorrefmark{1}}

\IEEEauthorblockA{\IEEEauthorrefmark{1}\textit{Fraunhofer Institute for Cognitive Systems IKS}\\
Garching, Germany \\
\{benjamin.herd, jessica.kelly\}@iks.fraunhofer.de}

\IEEEauthorblockA{\IEEEauthorrefmark{2}\textit{Technical University of Munich}\\
Garching, Germany \\
mario.trapp@tum.de}
}

%
%
%
\maketitle
%
\begin{abstract}
Machine learning (ML) is a key technology driving innovation today, but ensuring ML safety remains a major challenge for safety-related applications. A promising idea is to build \emph{proven‑in‑use arguments} from field data, e.g.\ by running ML components (MLCs) in shadow mode or within safety envelopes so that their outputs can be monitored as `safe probes' without affecting safety. These probes can then be used to build a statistical argument about field performance in a Bayesian way. However, many Bayesian field-data approaches in safety engineering model failures as a simple Bernoulli (or binomial) process with a single global failure probability and i.i.d.\ trials, which is rarely adequate for MLCs whose performance depends strongly on context. Statistical evidence is also about coverage of relevant situations, including edge cases, and building a single integrated statistical model for the entire system is usually not feasible. To address these challenges, this paper introduces CUBICS, a context‑modular framework for per‑component, situation‑aware performance estimation of safety-relevant ML components. CUBICS partitions the operational design domain into situations and, for each safety-relevant component, defines a set of situation-specific assumptions and probabilistic guarantees that are represented and updated in a Bayesian manner using Subjective Logic (SL). By combining these guarantees with beliefs about how often each situation occurs, CUBICS derives an overall risk estimate for each component without requiring a monolithic system-level statistical model, and thus provides a building block for modular, field‑data‑based safety assurance.
\end{abstract}

\begin{IEEEkeywords}
continuous safety assurance, machine learning, safety contracts
\end{IEEEkeywords}

\section{Introduction}
The safety assurance of systems based on Machine Learning (ML) remains a paramount challenge that necessitates the development of innovative paradigms, such as continuous safety assurance. This often involves operating ML components (MLCs) in shadow mode, where they function without any safety-relevant impact on the system to collect in-field data and establish a foundation of statistical evidence. 

Bayesian statistics provide a useful approach here: by treating each probe as a \textit{success} or \textit{failure}, a prior over the failure probability can be updated as data accumulates, yielding both an estimated failure probability and a measure of uncertainty. In this paper, we use the term `performance' to refer to task-level ML metrics such as recall, and `reliability' to refer to the probability that a safety-relevant component behaves correctly (i.e.\ achieves adequate performance) in a given situation. In practice, this is often instantiated as a simple Beta-Bernoulli model with a single global failure probability. For ML components whose performance is strongly context-dependent, this global i.i.d.\ assumption may mask situation-specific insufficiencies. For instance, a vision-based system is more likely to fail in heavy rain than in clear conditions. 

To address this, we introduce CUBICS, a framework for per‑component, situation‑aware performance estimation of safety‑relevant MLCs that can serve as a building block for modular, field‑data‑based safety assurance. CUBICS partitions the operational design domain (ODD) into discrete situations defined by context dimensions such as weather or lighting conditions. For each safety‑relevant component and situation, CUBICS defines a set of situation‑specific assumptions and probabilistic guarantees about relevant failure modes (e.g. false‑negative detection). Within each situation we assume approximately stationary failure behaviour, so simple Bernoulli/Beta‑style updates remain valid; at operation time, new evidence only updates the guarantees of the situations in which it may have occurred. This makes explicit which situations are well covered by evidence and where uncertainty about the component’s behaviour remains high.

Even with situation‑based models, it can be argued that compiling sound, fully integrated statistical evidence for system‑level claims (such as the positive risk balance of an automated driving system) is infeasible and prone to modelling errors. Building and maintaining a system‑wide Bayesian network (BN) that captures all relevant dependencies is particularly difficult in practice. In current industrial practice, complete statistical evidence for an entire system is typically neither available nor required; instead, system safety cases combine quantitative arguments for selected components with qualitative reasoning for the rest. CUBICS therefore uses Subjective Logic (SL) as the primary calculus for representing and updating component‑level, situation‑specific statistical evidence using subjective opinions with explicit belief and uncertainty. 

Prior work \cite{herd2024deductive} has formalised safety contracts in SL and derived a corresponding assurance argument pattern that separates assumption sufficiency from system resilience. That work focused on a single, context-agnostic binary contract and static evidence. In this paper, we extend this approach towards explicit ODD‑based situation modelling, per‑component context‑modular contracts, and runtime updates. For each safety‑relevant component, CUBICS defines an assume–guarantee contract. The assumptions capture beliefs about the context in which the component operates (e.g.\ distributions over weather or lighting conditions), while the guarantees express probabilistic statements about the component’s safety‑relevant behaviour, conditioned on these situations. Both assumptions and guarantees are represented as SL opinions that can be updated over time. By combining assumptions about situations with their conditional guarantees, CUBICS derives (i) situation‑specific guarantees and (ii) a marginal, context‑weighted overall risk contribution for the component. In principle, contracts of upstream components (e.g.\ perception) can be used as evidence for assumptions of downstream components (e.g. planning), but each contract can rely on different types of arguments (SL‑based models, traditional safety analyses, or qualitative reasoning). In this paper, we instantiate and evaluate CUBICS for an individual component and the composition of contracts across multiple components is left as future work. This preserves component-level modularity over ODD situations, uses Bayesian statistical methods where they are most needed for ML components, and avoids a single, fragile system‑wide statistical model in favour of feasible, component‑specific models that can feed into the overall safety case. In particular, we make the following contributions:

\begin{itemize}
  \item We introduce CUBICS, a \textbf{contract-based methodology for safety-relevant ML components} that structures assumptions as SL opinions over context dimensions and guarantees as SL opinions over situation-conditional failure behaviour, and derives both per-situation guarantees and a marginal, context-weighted risk contribution for an individual component.
  \item We develop a \textbf{context‑aware update mechanism} for CUBICS contracts that allows for the distribution of (positive and negative) runtime evidence across situations under context uncertainty. 
\end{itemize}

To investigate the effectiveness of our approach, we assess the following three research questions:

\begin{enumerate}
    \item \textbf{RQ1} (Internal validity):  Can CUBICS recover known, situation-specific reliability patterns in a controlled scenario?
    \item \textbf{RQ2} (Benefit): Does CUBICS yield more informative and situation-specific reliability assessments compared to a pooled Bernoulli model, and how is this advantage impacted by varying data sizes?
    \item \textbf{RQ3} (Sensitivity): How sensitive are the guarantees to prior choices, misclassification of context, and data scarcity?
\end{enumerate}

In the evaluation, we first instantiate CUBICS in a synthetic case study to assess its internal validity, i.e., whether it can recover known situation-specific reliability patterns and the corresponding marginal guarantee under controlled conditions (RQ1). We then apply CUBICS to a YOLOv12-based object detector trained on BDD100K, evaluating its behaviour on several safety-critical object classes (person, bicycle, car, etc.) and, due to space, report detailed results for the person class. We compare its situation-specific SL guarantees with a single global Bernoulli model to show that CUBICS exposes localized performance deficits and data gaps that the global model masks (RQ2). Finally, we analyse how the resulting guarantees change under different priors, imperfect context information, and varying evidence strength to assess the sensitivity of the approach to these factors (RQ3).

The paper is structured as follows. Section~\ref{sec:background} introduces Subjective Logic; Section~\ref{sec:cubics} presents the CUBICS methodology, including the contract model, situation‑based ODD decomposition, and the derivation of conditional and marginal guarantees with continuous updates; Sections~\ref{sec:exp}--\ref{sec:evaluation} provide experimental results on the application of CUBICS to an ML‑based safety‑relevant component and demonstrate the resulting context‑aware risk assessment; Section~\ref{sec:threats} discusses threats to validity; Section~\ref{sec:related_work} reviews related work; Section~\ref{sec:conclusion} summarises findings and outlines avenues for future work.

\section{Background}\label{sec:background}

\subsection{Subjective Logic}
\label{sec:subjective-logic}

Subjective Logic (SL)~\cite{josang2016subjective} is a framework for reasoning under uncertainty that combines ideas from probability theory and Dempster--Shafer evidence theory. Its core data structures are \emph{subjective opinions} representing an agent's belief, disbelief, and uncertainty about the truth of a proposition. SL provides algebraic operators for combining and transforming opinions. Depending on whether the underlying domain $\mathbb{X}$ is binary (i.e., $\mathbb{X} = \{x,\bar{x}\}$) or n-ary (i.e., $\mathbb{X} = \{x_1,\dots,x_K\}$), opinions are \textit{binomial} or \textit{multinomial}. We focus here on multinomial opinions.  

\begin{definition}[Multinomial opinion]
\label{def:multiop}
Let $\mathbb{X} = \{x_1,\dots,x_K\}$ be a finite domain of mutually exclusive and collectively exhaustive states.  
A \emph{multinomial opinion} over $\mathbb{X}$ is a tuple $\omega_{X} = (\mathbf{b}_X, u_X, \mathbf{a}_X)$ where:
\begin{itemize}
  \item $\mathbf{b}_X = (b_{x_1},\dots,b_{x_K})$ (\emph{belief masses}) is a distribution of belief over the states, with $b_{x_i}$ the belief mass supporting $x_i$ being the true state;
  \item $u_X$ (\emph{uncertainty}) is the remaining, uncommitted belief mass and the complement of \emph{confidence} ($1-u$);
  \item $\mathbf{a}_X = (a_{x_1},\dots,a_{x_K})$ (\emph{base rates}) is an a priori probability distribution over $\mathbb{X}$ in the absence of committed belief; and 
  \item $b_{x_i},u,a_{x_i} \in [0,1]$ for all $i$, $\sum_{i=1}^K b_{x_i} + u = 1$, and $\sum_{i=1}^K a_{x_i} = 1$.
\end{itemize}
\end{definition}
A \emph{vacuous} opinion (full uncertainty) is denoted $\omega_V = (\mathbf{0}, 1, \mathbf{a})$ and an \emph{absolute} opinion focusing all belief on a single state $x_i$ is denoted $\omega_{x_i}^\top = (\mathbf{b}, 0, \mathbf{a})$ with $b_{x_i}=1$ and $b_{x_j}=0$ for all $j \neq i$, for any choice of base rate vector $\mathbf{a}$.

\subsubsection{Constructing Multinomial opinions}
Given a finite domain $\mathbb{X} = \{x_1,\dots,x_K\}$, evidence counts 
$\mathbf{r} = (r_{x_1},\dots,r_{x_K})$ with $r_{x_i} \ge 0$ for each state $x_i$, and a 
\textit{non-informative prior weight}\footnote{$W$ ensures that, as evidence accumulates 
(i.e.\ $\sum_i r_{x_i}$ grows), the uncertainty $u$ decreases accordingly. $W$ is typically 
set to the cardinality of the domain ($W=K$), which is equivalent to adding one 
pseudo-observation to each state. Larger values of $W$ require more evidence for 
uncertainty to decrease.} $W$, a multinomial opinion can be computed as follows:
\begin{align}
    b_{x_i} &= \frac{r_{x_i}}{\sum_{j=1}^K r_{x_j} + W}, \quad i = 1,\dots,K
    \label{eq:multi-belief}\\[4pt]
    u &= \frac{W}{\sum_{j=1}^K r_{x_j} + W}
    \label{eq:multi-uncertainty}
\end{align}
with base rate vector $\mathbf{a} = (a_{x_1},\dots,a_{x_K})$, where 
$a_{x_i} \in [0,1]$ and $\sum_{i=1}^K a_{x_i} = 1$.

Multinomial opinions correspond to Dirichlet distributions over the categorical 
probabilities on $\mathbb{X}$. Given $(\mathbf{r}, \mathbf{a}, W)$, the corresponding 
parameters are
\begin{align}
    \alpha_{x_i} &= r_{x_i} + a_{x_i} W, \quad i = 1,\dots,K,
\end{align}
and the expectation value of $x_i$ is
\begin{align}
    E(x_i) = b_{x_i} + a_{x_i} \cdot u\label{eq:proj}
\end{align}

\subsubsection{Combining opinions} SL provides a wide range of combination operators \cite{josang2016subjective}. Combining opinions provides an elegant and intuitive way to combine the underlying distributions, a direct manipulation of which would be significantly more complex. In this paper, we use the following operators (see \cite{josang2016subjective} for full definitions):

\paragraph{Multinomial multiplication} given independent opinions $\omega_X$ and $\omega_Y$ about variables $X$ and $Y$ which take their values from distinct domains $\mathbb{X}$ and $\mathbb{Y}$, the joint opinion on the Cartesian product  $\mathbb{X} \times \mathbb{Y}$ is computed using multinomial multiplication $\omega_{X \wedge Y} = \omega_X \cdot \omega_Y$. We denote the corresponding joint belief as $b_{xy}$, uncertainty as $u_{xy}$, and product base rate as $a_{xy}$. Further details on computing multinomial multiplication can  be found in \cite{josang2016subjective}. 

\paragraph{Multinomial deduction} given a conditional relationship where conclusion variable $Y$ depends on premise variable $X$, the deduction operator derives the marginal opinion on $Y$ from an opinion on $X$ as $\omega_{Y\|X} = \omega_X \circledcirc \omega_{Y \mid X}$,where:
\begin{itemize}
    \item $\omega_X$ is the multinomial opinion on the premise $X$.
    \item $\omega_{Y \mid X}$ represents the set of conditional opinions on $Y$ given the mutually exclusive states of $X$.
    \item $\omega_{Y\|X}$ is the resulting deduced marginal opinion on $Y$.
    \item $\circledcirc$ denotes the deduction operator.
\end{itemize}
Full details for computing the marginal opinion $\omega_{Y\|X}$ are provided in \cite{josang2016subjective}. 

\section{The CUBICS methodology}\label{sec:cubics}

CUBICS structures the safety assurance of ML-based components (MLCs) around three central ideas:

\begin{enumerate}
    \item a situation-based decomposition of the operational design domain (ODD) is performed;
    \item modular safety contracts are defined per component and situation;
    \item an SL-based approach is used (1) to update beliefs about the current situation and the safety of the component based on operation-time evidence, and (2) to obtain conditional and marginal component safety guarantees.
\end{enumerate}

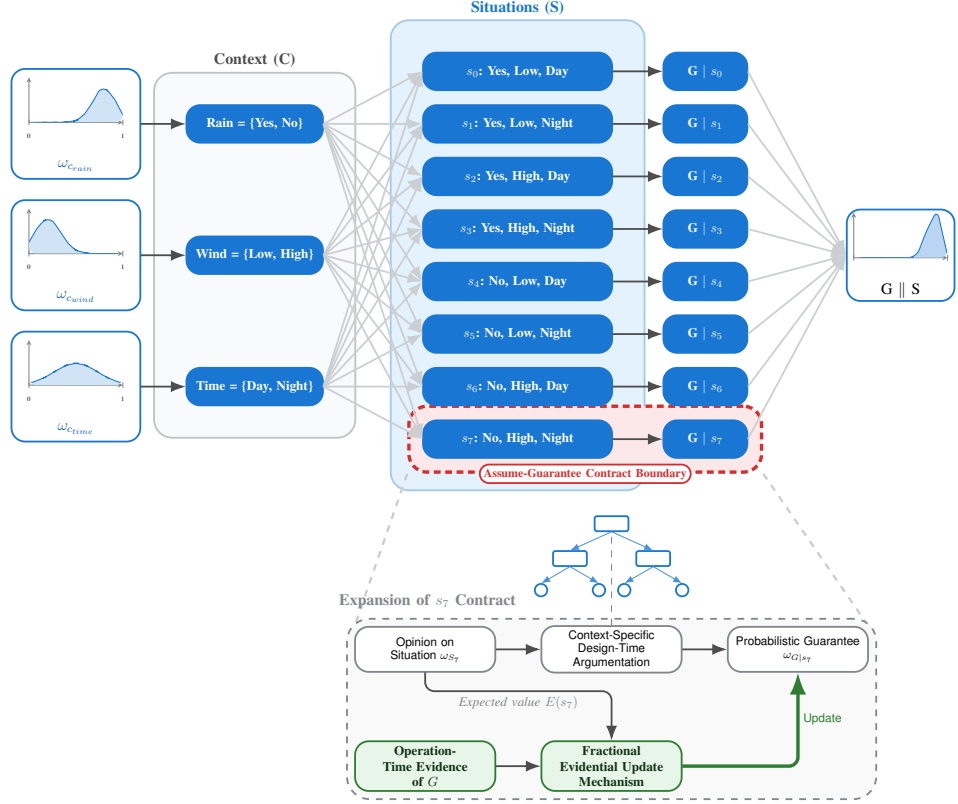
\begin{figure*}[!h]
\centering
\resizebox{0.7\textwidth}{!}{
\begin{tikzpicture}[
    scale=0.58,
    transform shape,
    font=\small\sffamily,
    >=Latex,
    box/.style={draw=myblue, thick, rounded corners=6pt, fill=myblue, text=white, minimum height=1.0cm, text width=3.4cm, align=center, font=\bfseries},
    state/.style={draw=myblue, thick, rounded corners=6pt, fill=myblue, text=white, minimum height=1.0cm, text width=4.8cm, align=center, font=\bfseries},
    gbox/.style={draw=myblue, thick, rounded corners=6pt, fill=myblue, text=white, minimum height=1.0cm, text width=2.0cm, align=center, font=\bfseries},
    final/.style={draw=myblue, thick, rounded corners=12pt, fill=myblue, text=white, minimum width=2.8cm, minimum height=1.4cm, align=center, font=\large\bfseries},
    logic/.style={draw=mygray, thick, rounded corners=6pt, fill=white, text=black, align=center, minimum height=1.2cm, text width=3.5cm},
    runtime/.style={draw=mygreen, thick, rounded corners=6pt, fill=mygreenbg, text=mygreen!40!black, align=center, minimum height=1.2cm, text width=3.5cm, font=\bfseries},
    panelC/.style={draw=mygray!50, thick, rounded corners=8pt, fill=mypanel, inner sep=14pt},
    panelS/.style={draw=myblue!40, thick, rounded corners=8pt, fill=mybluebg, inner sep=14pt},
    contract/.style={draw=myred, line width=1.5pt, densely dashed, rounded corners=8pt, fill=myredbg},
    flow/.style={->, thick, draw=black!70, rounded corners=4pt}, 
    flowcross/.style={->, thick, draw=mygray!40, rounded corners=4pt},
    updatearrow/.style={->, line width=1.5pt, draw=mygreen, rounded corners=4pt},
    pdfbox/.style={draw=myblue, thick, rounded corners=6pt, fill=white, inner sep=6pt, align=center, font=\large}
]


\node[box] (rain) at (0, 3.5) {Rain = \{Yes, No\}};
\node[box] (wind) at (0, 0) {Wind = \{Low, High\}};
\node[box] (time) at (0, -3.5) {Time = \{Day, Night\}};

\def\timevar{time}
\foreach \n/\p/\mu/\sig in {rain/p1/0.8/40, wind/p2/0.2/30, time/p3/0.5/10} {
    \node[pdfbox, left=1.2cm of \n] (\p) {
        \begin{tikzpicture}
            \begin{axis}[width=2.5cm, height=1.2cm, axis lines=left, axis line style={mygray, thick}, 
                xtick={0,1}, xticklabels={0,1}, ytick=\empty, tick label style={font=\tiny},
                xmin=0, xmax=1, ymin=0, ymax=1.2, enlarge x limits=false, scale only axis]
                \def\ampl{1.0} \ifx\n\timevar\def\ampl{0.6}\fi
                \addplot[name path=f, myblue, thick, domain=0:1, samples=10] {\ampl*exp(-\sig*(x-\mu)^2)};
                \path[name path=a] (axis cs:0,0) -- (axis cs:1,0);
                \addplot[fill=myblue!20] fill between[of=f and a];
            \end{axis}
        \end{tikzpicture}\\[2pt] \textcolor{myblue!80!black}{$\omega_{c_{\n}}$}
    };
}

\foreach \i/\txt in {0/{Yes, Low, Day}, 1/{Yes, Low, Night}, 2/{Yes, High, Day}, 3/{Yes, High, Night}, 4/{No, Low, Day}, 5/{No, Low, Night}, 6/{No, High, Day}, 7/{No, High, Night}} {
    \node[state] (s\i) at (7.0, {4.9 - \i*1.4}) {$s_{\i}$: \txt};
}

\foreach \i in {0,...,7} { \node[gbox] (g\i) at (12.0, {4.9 - \i*1.4}) {G $\mid$ $s_\i$}; }

\node[pdfbox, inner sep=2pt] (global) at (17.2, 0) {
    \begin{tikzpicture}
        \begin{axis}[
            width=2.5cm, height=1.2cm, 
            axis lines=left, 
            axis line style={mygray, thick}, 
            xtick={0,1}, xticklabels={0,1}, ytick=\empty, 
            tick label style={font=\tiny, color=white},
            xmin=0, xmax=1, ymin=0, ymax=1.2, 
            enlarge x limits=false, 
            scale only axis
        ]
            \addplot[name path=f_global, myblue, thick, domain=0:1, samples=10] {1.6*exp(-150*(x-0.85)^2)};
            
            \path[name path=a_global] (axis cs:0,0) -- (axis cs:1,0);
            
            \addplot[fill=myblue!30] fill between[of=f_global and a_global];
        \end{axis}
    \end{tikzpicture}\\[4pt]
    G $\parallel$ S
};


\coordinate (loopCenter) at (9.5, -10.5);

\node[logic] (argument) at (loopCenter) {Context-Specific\\Design-Time\\Argumentation};

\node[logic] (assump_logic) [left=1.2cm of argument] {
    Opinion on Situation $\opinion{S_7}$\\
};

\node[logic] (prob_guar) [right=1.2cm of argument] {
    Probabilistic Guarantee\\
    $\opinion{G \mid s_7}$ \\
};


\tikzset{
  gsnClaim/.style={
    draw=myblue,
    thick,
    rounded corners=1pt,
    fill=white,
    minimum width=0.9cm,
    minimum height=0.4cm
  },
  gsnEvidence/.style={
    draw=myblue,
    thick,
    circle,
    fill=white,
    minimum size=0.25cm
  },
  gsnArrow/.style={-latex, thin, draw=myblue!80},
  gsnRef/.style={dashed, thin, draw=mygray}
}

\node[gsnClaim] (gsnRoot) [above=2.5cm of argument] {};

\node[gsnClaim] (gsnSub1) [below left=0.5cm and 0.18cm of gsnRoot] {};
\node[gsnClaim] (gsnSub2) [below right=0.5cm and 0.18cm of gsnRoot] {};

\node[gsnEvidence] (gsnE1) [below left=0.5cm and 0.18cm of gsnSub1] {};
\node[gsnEvidence] (gsnE2) [below left=0.5cm and 0.18cm of gsnSub2] {};
\node[gsnEvidence] (gsnE3) [below right=0.5cm and 0.18cm of gsnSub1] {};
\node[gsnEvidence] (gsnE4) [below right=0.5cm and 0.18cm of gsnSub2] {};

\draw[gsnArrow] (gsnRoot.south) -- (gsnSub1.north);
\draw[gsnArrow] (gsnRoot.south) -- (gsnSub2.north);
\draw[gsnArrow] (gsnSub1.south) -- (gsnE1.north);
\draw[gsnArrow] (gsnSub2.south) -- (gsnE2.north);
\draw[gsnArrow] (gsnSub1.south) -- (gsnE3.north);
\draw[gsnArrow] (gsnSub2.south) -- (gsnE4.north);

\draw[gsnRef] (argument.north) -- (gsnRoot.south);

\node[runtime, below=1.8cm of argument] (update) {
    Fractional Evidential Update\\
    Mechanism
};

\node[runtime, left=1.2cm of update] (evidence) {
    Operation-Time Evidence\\
    of $G$
};


\draw[flow] (assump_logic) -- (argument);
\draw[flow] (argument) -- (prob_guar);

\draw[flow] (assump_logic.south) -- ++(0,-0.5) -| node[pos=0.25, below, font=\itshape, text=mygray] {Expected value $E(s_7)$} (update.north);

\draw[flow] (evidence) -- (update);

\draw[updatearrow] (update.east) -| node[pos=0.75, right, text=mygreen, align=left] {Update} (prob_guar.south);

\begin{scope}[on background layer]
    \node[panelC, fit=(rain) (time), label={[font=\large\bfseries, text=black!70]above:Context (C)}] (cpanel) {};
    \node[panelS, fit=(s0) (s7), label={[font=\large\bfseries, text=myblue]above:Situations (S)}] (spanel) {};
    
    \node[contract, inner sep=6pt, fit=(s7.north west) (g7.south east)] (contractbox) {};

    \draw[dashed, gray!40, line width=1pt] (contractbox.south west) -- (assump_logic.north west);
    \draw[dashed, gray!40, line width=1pt] (contractbox.south east) -- (prob_guar.north east);
    
    \node[draw=mygray, dashed, thick, rounded corners=10pt, fill=mygray!5, 
          fit=(assump_logic) (prob_guar) (evidence), 
          label={[font=\large\bfseries, text=mygray, yshift=4pt]above left:Expansion of $s_7$ Contract}] (runtime_panel) {};

    \node[font=\small\bfseries, text=myred, fill=white, draw=myred, thick, rounded corners=4pt, inner sep=3pt] 
        at (contractbox.south) {Assume-Guarantee Contract Boundary};
\end{scope}

\draw[flow] (p1.east) -- (rain.west); 
\draw[flow] (p2.east) -- (wind.west); 
\draw[flow] (p3.east) -- (time.west);

\foreach \dst in {0,...,7} {
    \draw[flowcross] (rain.east) -- (s\dst.west); 
    \draw[flowcross] (wind.east) -- (s\dst.west); 
    \draw[flowcross] (time.east) -- (s\dst.west);
}
\foreach \i in {0,...,7} { 
    \draw[flow] (s\i.east) -- (g\i.west); 
    \draw[flowcross] (g\i.east) -- (global.west); 
}

\end{tikzpicture}}
\caption{Overview of the Unified CUBICS methodology over a simplified operational design domain. Subjective context opinions over \emph{Rain}, \emph{Wind}, and \emph{Time of Day} make up eight situations (from $s_0$ to $s_7$). Each situation establishes a unique safety contract ($G\mid s_i$), which are ultimately aggregated into the overarching global guarantee $G\parallel S$.}
\label{fig:cubics_workflow_example}
\end{figure*}

This section explains how these ideas jointly yield a context-aware assessment of a component's safety guarantees as well as the overall risk contribution that can be embedded into a broader safety case. Throughout this section, we refer to the simplified scenario in Fig.~\ref{fig:cubics_workflow_example} as a running example.

\begin{figure*}[!htbp]
\centering
\resizebox{0.7\textwidth}{!}{
\begin{tikzpicture}[
    scale=0.85,
    transform shape,
    font=\small\sffamily,
    >=Latex,
    box/.style={draw=myblue, thick, rounded corners=6pt, fill=myblue, text=white, minimum height=1cm, text width=3.2cm, align=center, font=\bfseries},
    logic/.style={draw=mygray, thick, rounded corners=6pt, fill=white, text=black, align=center, minimum height=1cm, text width=3.2cm},
    runtime/.style={draw=mygreen, thick, rounded corners=6pt, fill=mygreenbg, text=mygreen!40!black, align=center, minimum height=1cm, text width=3.8cm, font=\bfseries},
    math/.style={draw=mygray!40, fill=mygray!5, rounded corners=4pt, inner sep=6pt, font=\footnotesize\ttfamily, align=left},
    flow/.style={->, thick, draw=black!70, rounded corners=4pt},
    updatearrow/.style={->, line width=1.5pt, draw=mygreen, rounded corners=4pt},
    labelnode/.style={font=\tiny\itshape, text=mygray}
]

\node[box] (sit_opinion) at (0, 3) {Situation Opinion\\$\opinion{s_7}$};
\node[runtime] (field_data) at (0, 0) {Field Evidence Stream\\$x \in \{0, 1\}$};

\node[runtime] (weight) at (5, 3) {Weighting Function\\$E(s_7) = b_{s_7} + a_{s_7}u_{s_7}$};

\node[runtime] (fractional) at (5, 0) {Fractional Evidence\\Accumulation};

\node[math] (formula) at (5, -1.8) {
    $\Delta r = E(s_7) \cdot \mathbb{I}(x=1)$\\
    $\Delta s = E(s_7) \cdot \mathbb{I}(x=0)$
};

\node[logic] (dirichlet) at (10, 1.5) {Refined Guarantees};

\node[box, right=1.5cm of dirichlet] (final_g) {Updated Conditional Guarantee\\$\opinion{G \mid s_7}$};

\node[below=0.2cm of final_g] (miniplot) {
    \begin{tikzpicture}
        \begin{axis}[width=2.5cm, height=1.2cm, axis lines=left, ticks=none, xmin=0, xmax=1, ymin=0, ymax=1.2, enlarge x limits=false, scale only axis]
            \addplot[name path=old, mygray!40, dashed, domain=0:1, samples=30] {exp(-10*(x-0.5)^2)}; 
            \addplot[name path=new, myblue, thick, domain=0:1, samples=50] {exp(-80*(x-0.7)^2)}; 
            \path[name path=ax] (axis cs:0,0) -- (axis cs:1,0);
            \addplot[fill=myblue!20] fill between[of=new and ax];
        \end{axis}
    \end{tikzpicture}
};


\draw[flow] (sit_opinion) -- (weight);
\draw[flow] (weight) -- node[labelnode, right] {Weight $P(s_7)$} (fractional);
\draw[flow] (field_data) -- node[labelnode, above] {} (fractional);
\draw[flow, dashed] (formula) -- (fractional);

\draw[updatearrow] (fractional.east) -| ++(0.6, 1.5) -- (dirichlet.west);
\draw[updatearrow] (dirichlet.east) -- node[above, text=mygreen, font=\bfseries\tiny] {} (final_g.west);

\begin{scope}[on background layer]
    \node[draw=mygreen!40, dashed, fill=mygreen!5, rounded corners=12pt, fit=(weight) (fractional) (formula), 
          label={[text=mygreen!60!black, font=\bfseries]above:Fractional Update Mechanism}] (panel) {};
\end{scope}

\end{tikzpicture}}
\caption{The fractional evidential update mechanism. SL situational opinions $\opinion{s_i}$ provide a probabilistic weighting signal $E(s_i)$ to incoming field observations. Binary evidence is accumulated fractionally into Dirichlet parameters $(\alpha, \beta)$, thereby refining the conditional guarantee $\opinion{G \mid s_i}$ as more data is gathered.}
\label{fig:fractional-update}
\end{figure*}
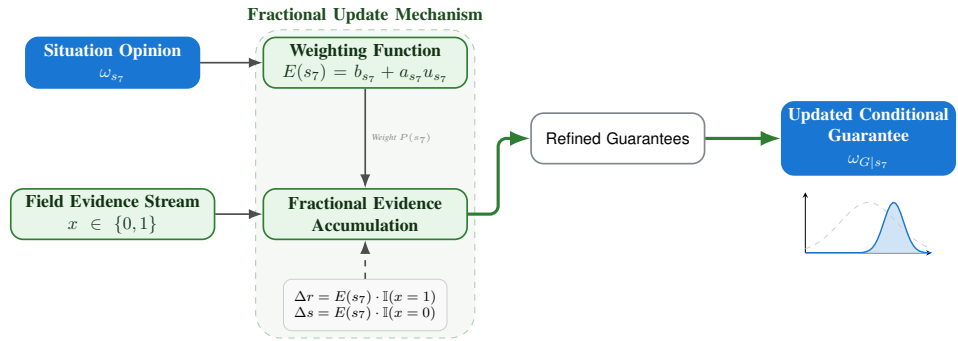

\subsection{Situation-based decomposition of the ODD}
\label{sec:odd-decomposition}

We assume that the relevant ODD of the system
can be characterised by a set of discrete \textit{context dimensions} $C_1, \ldots, C_k$ (e.g.\ weather, lighting). Each dimension $C_i$ is a finite set of possible values, e.g.\ $C_1\!=\!\mathit{Rain}\!=\!\{\mathit{Yes}, \mathit{No}\}$ and $C_2\!=\!\mathit{Wind}\!=\!\{\mathit{Low}, \mathit{High}\}$. CUBICS partitions the ODD into \emph{situations} $s \in S$, where $S\!=\!C_1 \times C_2 \times \cdots \times C_k$, and each situation $s = (c_1,\ldots,c_k) \in S$ corresponds to a particular combination of context values (e.g.\ $\mathit{Rain}\!=\!\mathit{Yes}$,$\mathit{Wind}\!=\!\mathit{High}$). Each such situation represents a subspace of the ODD that is assumed to have a distinct performance or risk profile for the component under consideration. For example, with $\mathit{Rain}\!=\!\{\mathit{Yes}, \mathit{No}\}$ and $\mathit{Wind}\!=\!\{\mathit{Low}, \mathit{High}\}$ we obtain four situations, and adding a third dimension $\mathit{Time}\!=\!\{\mathit{Day}, \mathit{Night}\}$ yields the eight situations shown in Fig.~\ref{fig:cubics_workflow_example}.

The key assumption is that, in each situation $s_i$, the component's failure behaviour can be treated as approximately stationary, so that a Bernoulli model of success and failure is more representative. More precisely, individual outcomes of the considered failure mode (e.g.\ misclassifications of an object) are exchangeable Bernoulli trials with an (approximately) constant failure probability $p_{fail}(s_i)$. In the contract, the resulting context model (assumptions over $C_i$ and $S$) is linked to per-situation guarantees $\opinion{G \mid s_i}$ as introduced below.

\subsection{Contracts for safety-relevant components}\label{sec:contracts}

For each MLC and situation $s_i$ as defined above, CUBICS defines a \textit{safety contract} with two central elements:

\begin{itemize}
  \item \textbf{Assumptions} capture probabilistic beliefs about the current context, i.e.\ which values the context dimensions take, e.g.\ $Rain{=}\{\mathit{Yes},\mathit{No}\}$, $Wind{=}\{\mathit{Low},\mathit{High}\}$, and thus which situation $s_i$ the component is believed to be operating in. These beliefs result from environment perception, operational profiles, or scenario-based analyses. As described below, beliefs are represented by multinomial opinions in SL which are then combined into a joint situational assessment.
  \item \textbf{Guarantees} express beliefs about the safety of the component or function, \emph{conditional on the current situation}. They are underpinned by a safety case with appropriate evidence, such as design-time analyses, field or shadow-mode data, and expert judgement. For each situation $s_i$, belief in the guarantee is represented as a probabilistic assessment of the relevant safety claim $G$, encoded as a conditional opinion $\opinion{G \mid s_i}$.
\end{itemize}

Assumptions and guarantees are the building blocks of a conditional model that links beliefs about operating in a certain situation $s_i$ with beliefs about the safety of the component in $s_i$ based on available evidence. Conceptually, this conditional model resembles a Bayesian Network (BN) but replaces conditional probability tables with conditional probability density functions. More precisely, CUBICS uses SL to represent assumptions and guarantees as follows:

\begin{itemize}
    \item Assumptions are modelled by first forming multinomial opinions $\opinion{C_i}$ about each context dimension $C_i$; these opinions are then combined into a joint situational opinion $\opinion{S} = \prod_{i=1}^k \opinion{C_i}$.
    \item Guarantees are situation-specific and are thus represented as a set $\opset{G\mid S} = \{ \,\opinion{G \mid s_i} \mid s_i \in \mathit{S} \,\}$ of conditional opinions. Conceptually, $\opset{G \mid S}$ forms a context-indexed family of opinions, one per situation $s_i$. 
\end{itemize}

Using the example from Fig.~\ref{fig:cubics_workflow_example}, a belief such as \emph{``it is likely clear, not windy, and daytime''} is first encoded in the individual context opinions (e.g.\ $\omega_{\mathit{Rain}}$, $\omega_{\mathit{Wind}}$, $\omega_{\mathit{Time}}$) and then combined into the joint situational opinion $\omega_S$, which in turn links directly to the corresponding per-situation guarantee, e.g.\ $\omega_{G \mid s_4}$ for $(\mathit{Rain}=\mathit{No}, \mathit{Wind}=\mathit{Low}, \mathit{Time}=\mathit{Day})$.

The approach is component‑level and context‑modular: in a system with multiple interacting components, each component is equipped with its own CUBICS contract over ODD situations. In principle, the contracts of `upstream' components (e.g.\ perception) can provide SL opinions over their guarantees that can be used as evidence for the assumptions of `downstream' contracts (e.g.\ planning or actuation). In this way, guarantees do not need to be recomputed within a single monolithic probabilistic model: each component maintains its own context-dependent guarantees in the SL space, and the system-level safety case connects these contracts by treating upstream guarantees as inputs to downstream assumptions.  This avoids a single, fragile system‑wide statistical model in favour of component‑specific models that can feed into the overall safety case. Based on this conditional model, both situation-specific guarantees and an overall risk contribution can be derived, as described below. Note that, in this paper, we instantiate and evaluate CUBICS at the level of individual components; the use of composed contracts across multiple components is left for future work.

\subsection{Conditional and marginal contract guarantees}\label{sec:guarantees}

A component contract gives rise to two closely related types of guarantees: the \emph{conditional} view maintains one guarantee per situation -- explicitly conditioned on $s_i \in S$ -- while the \emph{marginal} view aggregates these per-situation guarantees into a single overall guarantee under the modelled operational profile. We describe both views below.

\subsubsection{Conditional view (one guarantee per situation)}

As defined in Section \ref{sec:contracts}, for each situation $s_i \in S$ we maintain a conditional opinion $\opinion{G \mid s_i}$ over a binary guarantee domain $G$ (e.g. ${Safe, Unsafe}$). The collection of these opinions, $\opset{G\mid S} = \{ \,\opinion{G \mid s_i} \mid s_i \in \mathit{S} \,\}$, constitutes the conditional view of the contract. This view supports detailed, context-specific reasoning about the component in each situation.

\subsubsection{Marginal view (one aggregated guarantee over all situations)}

In contrast, the \emph{marginal} view considers the contract from the perspective of the actual operation of the system, where different situations occur with different
probabilities according to the operational profile and context model. Here we are interested in a \emph{single} overall opinion about the guarantee $G$ that already takes
into account how likely each situation is.

CUBICS obtains this overall opinion by combining: (i) the SL opinions over the context dimensions (assumptions), which induce a belief over situations $s_i$, and
(ii) the conditional per-situation opinions $\opinion{G \mid s_i}$. This combination is performed by multinomial deduction in SL, which yields a single marginal opinion $\opinion{G||S}$ which, for simplicity, we abbreviate with $\opinion{G}$:
\begin{align*}
  \opinion{G} \coloneq \opinion{G||S} = \opinion{S} \deduction \opset{G|S}
\end{align*}
where $\opinion{S} = \prod_{i=0}^k \opinion{C_i}$ denotes the joint opinion over all possible context dimensions $C_i$, computed via multinomial conjunction.\footnote{In practice, this corresponds to combining the SL opinions over the context variables into a joint belief over situations.} Intuitively, multinomial deduction weighs each per-situation guarantee $\opinion{G \mid s_i}$ by the belief that the corresponding situation $s_i$ actually occurs, and aggregates the results into one overall conclusion. The marginal opinion $\opinion{G}$ can be understood as the component's \emph{overall risk contribution} under the modelled operational profile, i.e.\ as a probabilistic
assessment (with uncertainty) of whether its guarantee will be satisfied during operation when all situations and their probability of occurrence are taken into account.

In summary, the conditional view provides a set $\opset{G \mid S}$ of per-situation guarantees $\opinion{G \mid s_i}$, while the marginal view provides one aggregated opinion $\opinion{G}$.
Both are derived from the same contract: engineers can inspect per-situation guarantees for detailed, context-specific reasoning, and use the marginal guarantee as a compact summary of the component's overall risk contribution.

\subsection{Continuous update of situational beliefs and guarantees}\label{sec:continuous_update}

CUBICS follows a Bayesian perspective: an important design principle is that neither the situational opinions $\opinion{C_i}$ about the individual context dimensions nor the conditional guarantee opinions $\opinion{G \mid s_i}$ stated in a contract are static. As additional evidence becomes available, both sets of opinions are updated, which, in turn, influences both the situational joint opinion set $\opset{S}$ and the inferred marginal risk opinion $\opinion{G}$. CUBICS focuses on runtime updates from statistical evidence obtained during operation or testing. Here, operation-time data from field or shadow-mode deployments, as well as additional test datasets, are mapped into SL opinions and fused with the existing contextual and conditional per-situation opinions $\opinion{C_i}$ and $\opinion{G \mid s_i}$. In Bayesian terms, this corresponds to an update of the underlying Dirichlet parameters. Over time, this process refines beliefs and reduces uncertainty about both the occurrence probability of each situation $s_i$ as well as the guarantees that can be given in $s_i$. CUBICS uses a \textit{fractional update mechanism} to account for uncertainty in the assumptions about which situation we are in. Given opinion $\opinion{S} = \prod_{i=1}^k \opinion{C_i}$ derived from opinions $\omega_{C_i}$ on context parameters, we compute the expected probability $E(s_i)$ of being in situation $s_i$ according to Eq.~\ref{eq:proj} and perform an update of the conditional guarantees $\opinion{G \given s_i}$ by weighing observed evidence $x$ by the corresponding expectation value:
  \begin{align}
      r_i^{(t+1)} &= r_i^{(t)} + E(s_i)\,\mathbb{I}[x_t = 1], \\
      s_i^{(t+1)} &= s_i^{(t)} + E(s_i)\,\mathbb{I}[x_t = 0].
\end{align}
where $r_i$ and $s_i$ are the accumulated success and failure counts for situation $s_i$, and $\mathbb{I}[\cdot]$ is the indicator function. For example, if $\opinion{S}$ projects $E(s_0)=0.8$ and $E(s_1)=0.2$ at time $t$ and we observe a failure ($x_t = 0$), then the failure counts are updated by $s_0^{(t+1)} = s_0^{(t)} + 0.8$ and $s_1^{(t+1)} = s_1^{(t)} + 0.2$. An overview of the fractional update mechanism is provided in Figure \ref{fig:fractional-update}. 

By operating in the SL space, CUBICS allows practitioners to transparently integrate heterogeneous statistical evidence sources and domain knowledge, and to reflect both supporting and adverse observations in the evolving contract. The resulting per-situation guarantees and marginal risk assessments are continuously updated in a Bayesian way and may serve as an input to higher-level assurance reasoning without requiring a monolithic, system-wide statistical model.  
\section{Experimental Set-Up} \label{sec:exp}
In this section, we describe our experimental setup, including the Python implementation of CUBICS, architectural decisions, training pipeline details, and used datasets. The source code is available at \url{https://doi.org/10.5281/zenodo.21932463}. 

\subsubsection{CUBICS Python implementation}
The methodology has been implemented as a Python framework that provides executable support for defining context-dependent safety contracts, representing them in SL, and updating them continuously as new evidence becomes available. Conceptually, it mirrors the structure described in Section \ref{sec:cubics}. Context dimensions (e.g. weather, lighting, time of day) are represented as multinomial SL opinions and combined into a joint situational opinion over the Cartesian product of all context dimensions. For each resulting situation, the framework maintains a conditional opinion that captures the corresponding contract guarantees. These per-situation guarantees are linked to the situational opinion via the SL deduction operator, yielding a single marginal guarantee that summarises the component’s overall risk contribution under its operational profile. The python code also implements the fractional evidential update mechanism introduced in Section \ref{sec:continuous_update}. Incoming operation-time evidence is weighted by the expected value of the situations in which it may have occurred and accumulated into the corresponding per-situation guarantees. The implementation builds on standard Python scientific computing tools, facilitating integration into simulation environments and safety-assurance workflows.

\subsubsection{Datasets}
Experiments were conducted using the Berkeley DeepDrive (BDD100K) dataset \cite{BDD100K}. It was selected due to its high diversity in driving scenarios and the inclusion of frame-level environmental attributes. The dataset consists of 100,000 annotated images, divided into a 70k/10k/20k split for training, validation, and testing, respectively. To support our analysis, the standard 10k validation set was stratified into mutually exclusive sub-datasets based on the weather attribute present in the original JSON annotations. The evaluated conditions include: \textit{Clear}, \textit{Overcast}, \textit{Rainy}, \textit{Snowy}, and \textit{Foggy}. We pragmatically restrict our analysis to binary context dimensions, given the constraints of the existing BDD100K annotations. However, CUBICS itself is not limited to binary variables; because the SL opinions over context dimensions are multinomial, they can naturally represent multiple values per dimension. The models were trained to detect 10 standard classes: \textit{person}, \textit{rider}, \textit{car}, \textit{truck}, \textit{bus}, \textit{train}, \textit{motorcycle}, \textit{bicycle}, \textit{traffic light}, and \textit{traffic sign}.

\subsubsection{Model Architecture and Training}
We used the YOLOv12-Large (YOLOv12l) architecture \cite{tian2025yolov12} as our primary baseline, initialized with pre-trained COCO weights to accelerate training. To preserve the fidelity of small, distant objects (such as pedestrians in low-visibility conditions), the input image resolution was scaled to $1024 \times 1024$ pixels. Training was conducted over 100 epochs.  

\subsubsection{Hardware}
All experiments were executed on a single NVIDIA GeForce RTX 4090 GPU with 24GB of VRAM.

\section{Evaluation}\label{sec:evaluation}
In this section we provide quantitative results addressing RQ1, RQ2, and RQ3. To address RQ1, we provide a synthetic case study to highlight the CUBICS methodology. We then evaluate RQ2 and RQ3 on the experimental setup outlined in Section \ref{sec:exp}. 

\subsection{\textbf{RQ1 (Internal validity)}:  Can CUBICS recover known, situation-specific reliability patterns in a controlled scenario?}
To address RQ1, we use a synthetic scenario that mimics a perception component operating in an environment with known situation-specific failure rates. This controlled setup allows us to verify whether CUBICS can reconstruct the underlying per-situation reliability patterns\footnote{Here, by \textit{reliability patterns} we mean the mapping from each situation $s_i$ to its ground-truth probability of correct behaviour (e.g., classification accuracy) and the corresponding differences between situations (e.g., some situations being systematically harder than others).} and the resulting marginal guarantee from observed success/failure outcomes. We consider an abstract perception-based function (e.g.\ a pedestrian detector) in the context of automated driving whose failure behaviour is assumed to depend on three discrete context dimensions: $Rain{=}\{Yes,No\}$, $Wind{=}\{High,Low\}$, and $ToD{=}\{Day,Night\}$. The Cartesian product of these dimensions induces eight distinct situations, each of which is assumed to have an approximately stationary failure probability. For each situation, we define a conditional guarantee about the component’s safety-relevant behaviour and combine them with beliefs about the occurrence of the situations to obtain a marginal, context-weighted risk contribution. All steps of this process are implemented and executed in Python.

\paragraph{Design-time} Following the CUBICS methodology, the assessment of the ML component is represented as a safety contract that is evaluated against the situations induced by the context dimensions $Rain$, $Wind$, and $ToD$. The contract assumptions capture beliefs about the current context and are represented as multinomial SL opinions $\opinion{R}$, $\opinion{W}$, and $\opinion{T}$, which are then combined into a joint situational opinion $\omega_S$ over the eight situations $S = Rain \times Wind \times ToD$ using multinomial multiplication. The contract guarantees are concerned with a single safety-relevant property $G$ of the component (e.g.\ no safety-relevant miss occurs in a critical detection scenario) and are made explicit on a per-situation basis. At design time, engineers construct a safety case that argues, for each situation $s_i \in S$, for bounds on the probability that $G$ is satisfied when the component operates in $s_i$. The detailed argumentation (tests, analyses, expert judgement) is not modelled explicitly here; instead, its outcome is summarised as an SL opinion $\omega_{G \mid s_i}$ over the binary domain $G = \{Safe, Unsafe\}$. The collection of these eight conditional opinions $\omega_{G \mid S} = \{\omega_{G \mid s_i} \mid s_i \in S\}$ constitutes the conditional part of the contract and forms the design-time initialisation of the case study.

\begin{figure*}[t]
    \centering
    \includegraphics[width=\linewidth]{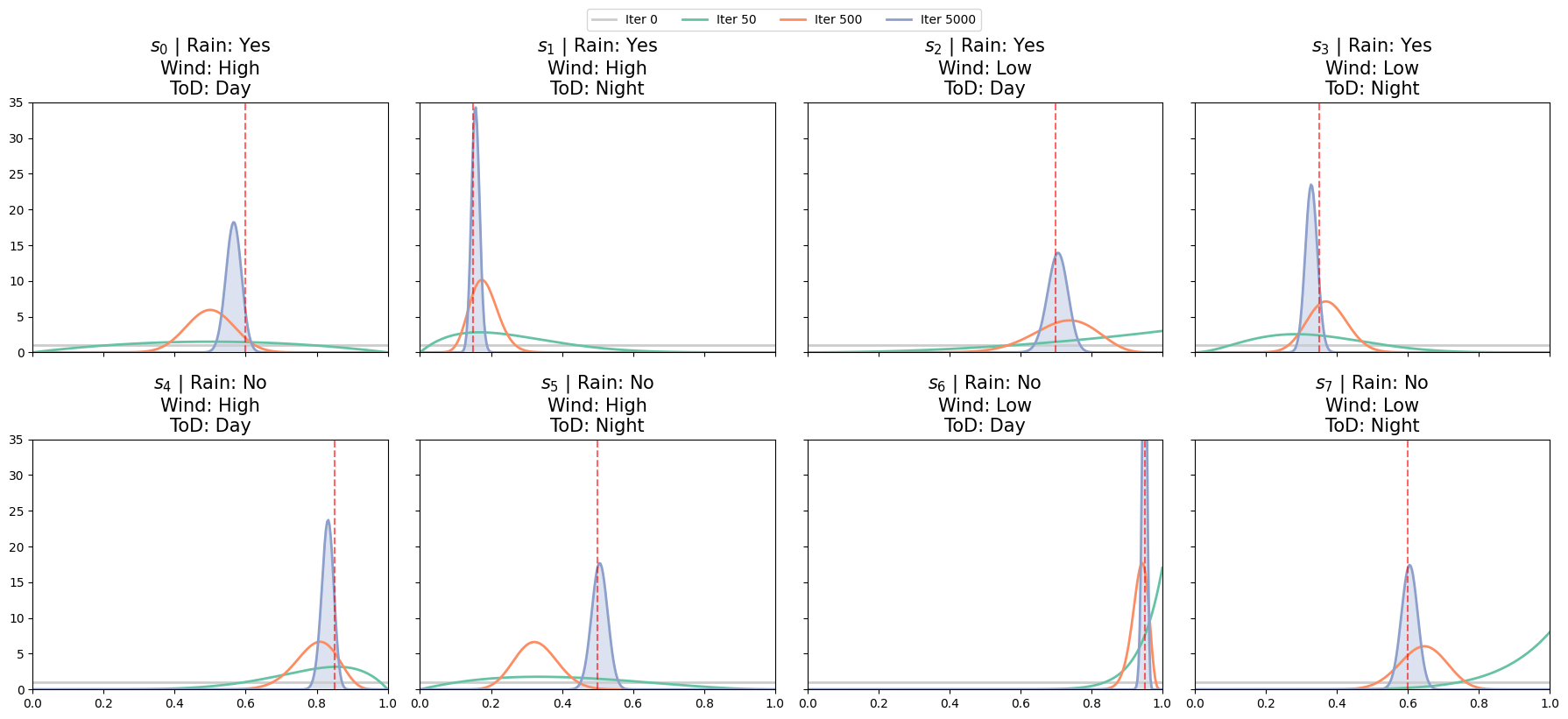}
    \caption{Progression of beta distributions across situational guarantees. Dotted red line represents true failure probability across situations. The shaded beta distributions represent the final conditional guarantees.}
    \label{fig:evolution}
\end{figure*}

\paragraph{Operation-time.} At operation-time, we initialize both the context opinions $\omega_R$, $\omega_W$, $\omega_T$ and the conditional guarantees $\omega_{G \mid S}$ with near-complete uncertainty (vacuous priors). This represents a system deployed in shadow mode with minimal prior empirical data, relying instead on operation-time evidence to learn the behaviour across all conditions. We simulate a deployment over $N = 5{,}000$ discrete operational cycles. Each situation $s_i$ is assigned a ground-truth physical failure rate $p_{\text{fail}}(s_i)$. These rates reflect intuitive physical constraints; for instance, the failure probability is highest in the most adverse conditions ($Rain = Yes$, $Wind = High$, $ToD = Night$) and lowest in ideal conditions. During each simulated cycle, a situation is sampled from the ground-truth context distribution. The agent records this occurrence and accumulates environmental evidence to continuously update the base context opinions $\omega_R, \omega_W$, and $\omega_T$. Simultaneously, the component's performance w.r.t.\ its guarantees is simulated against the target situation's $p_{\text{fail}}(s_i)$, yielding a binary success ($r$) or failure ($s$) outcome. This specific evidence is mapped to the corresponding conditional opinion $\omega_{G \mid s_i}$. In this synthetic setting, the true situation is observed without ambiguity, so the situational opinion $\omega_S$ collapses to a point mass (i.e.\ $P(s_i)=1$ for the realised situation and $0$ otherwise). Consequently, the fractional evidential update mechanism described in Section \ref{sec:continuous_update} reduces to a standard per-situation Bernoulli/Beta update; in scenarios with uncertain context, the same mechanism would distribute each observation fractionally across multiple situations according to $E(s_i)$.

\paragraph{Results}
Figure~\ref{fig:evolution}  
illustrates the evolution of the conditional safety guarantees ($\omega_{G \mid s_i}$) for all eight context situations over $N = 5{,}000$ operational cycles. At iteration 0, the beliefs are initialized with vacuous priors, represented by the flat, uniform distribution (grey line). As the system accumulates evidence, epistemic uncertainty decreases, and the Beta distributions progressively sharpen (iterations 50 and 500). By iteration 5,000, the expected value of every conditional opinion has converged towards its respective ground-truth success probability (red dashed line). The final distributions exhibit varying degrees of uncertainty (represented by the width and peak density of the curves), accurately reflecting the underlying occurrence probabilities $E(s_i)$ of the respective situation $s_i$. Frequently encountered situations (e.g., $s_1$ and $s_6$) accumulate substantial evidence, resulting in highly confident, narrow density peaks. Conversely, rare situations (e.g., $s_2$ or $s_5$) accumulate less operational evidence, naturally resulting in wider distributions. This demonstrates that the CUBICS framework correctly preserves uncertainty where data is scarce, preventing the system from making overconfident safety claims about rare edge cases.

\begin{RQSummary}{}
\textbf{RQ1 Summary}: In this controlled scenario, CUBICS converges towards the known situation-specific reliability patterns: per-situation guarantees $\opinion{G \mid s_i}$ recover the predefined failure rates within their uncertainty bounds. This indicates that, when its assumptions hold, CUBICS correctly captures situation-dependent reliability from observed success/failure outcomes.
\end{RQSummary}

\subsection{\textbf{RQ2 (Benefit)}: Does CUBICS yield more informative and situation-specific reliability assessments compared to a global pooled Bernoulli estimate, and how is this advantage impacted by varying available data sizes?}

To answer RQ2, we analyse the YOLOv12l object detector on the BDD100K validation set as described in Section~\ref{sec:exp}, and compare two different strategies for deriving a system-level recall guarantee from the same body of evidence for the \texttt{person} class. We repeated the analysis for other object classes and observed qualitatively similar behaviour; we therefore focus the following description primarily on the \texttt{person} class for brevity. Each ground-truth \texttt{person} instance detection is treated as a Bernoulli trial, so we use true positive (TP) and false negative (FN) counts as evidential successes and failures for the true-positive rate (recall):

\begin{enumerate}
    \item \textbf{Global pooled guarantee.} We consider a global pooled estimate for comparison, where all TP and FN counts are pooled into a single opinion $\omega_{\text{gp}}$ that yields one context-agnostic overall guarantee.
    \item \textbf{SL marginal guarantee.} The ODD is decomposed into $k$ situations. A conditional opinion $\omega_{G|s_i}$ is formed from the per-situation evidence, a multinomial context opinion $\omega_S$ captures the situation distribution, and the marginal $\omega_{G\|S}$ is obtained via SL deduction~\cite{josang2016subjective}. This also yields one global guarantee, but it preserves both evidential and contextual uncertainty.
\end{enumerate}

\paragraph{Situational guarantees expose masked performance insufficiencies.}
Table~\ref{tab:cubics_rq2} reports the per-situation opinions for the person class. The global pooled estimate concentrates at $E = 0.653$ with negligible uncertainty ($u \approx 10^{-4}$), suggesting high confidence in a moderate recall. CUBICS reveals that this figure conceals situation-specific performance insufficiencies: \emph{Clear\,+\,Daytime} achieves a comparable expected probability ($E = 0.650$), whereas \emph{Foggy\,+\,Dawn/dusk} collapses to $E = 0.250$ with uncertainty elevated by two orders of magnitude ($u = 0.083$). This adverse scenario contains only 22 observations, making the wide posterior an honest reflection of evidential scarcity.

\begin{table}[H]
    \centering
    \caption{CUBICS opinions for the \texttt{person} class. The global pooled estimate masks the performance collapse and high uncertainty in adverse scenarios such as \emph{Foggy\,+\,Dawn/dusk}.}
    \label{tab:cubics_rq2}
    \resizebox{\columnwidth}{!}{%
\begin{tabular}{@{}l r r c c c c@{}}
    \toprule
    \textbf{Scenario} & \textbf{TP ($r$)} & \textbf{FN ($s$)} & \textbf{Bel. ($b$)} & \textbf{Disbel. ($d$)} & \textbf{Unc. ($u$)} & \textbf{Exp.\ Prob ($E$)} \\
    \midrule
    Global Pooled & 16\,101 & 8\,549 & 0.6531 & 0.3468 & 0.0001 & 0.6532 \\
    Clear + Daytime   & 2\,758  & 1\,484 & 0.6499 & 0.3497 & 0.0005 & 0.6501 \\
    Rainy + Night     & 218     & 157    & 0.5782 & 0.4164 & 0.0053 & 0.5809 \\
    Foggy + Dawn/Dusk       & 5       & 17     & 0.2083 & 0.7083 & 0.0833 & 0.2500 \\
    \bottomrule
\end{tabular}%
    }
\end{table}

\begin{figure}[t]
    \centering
    \includegraphics[width=1\columnwidth]{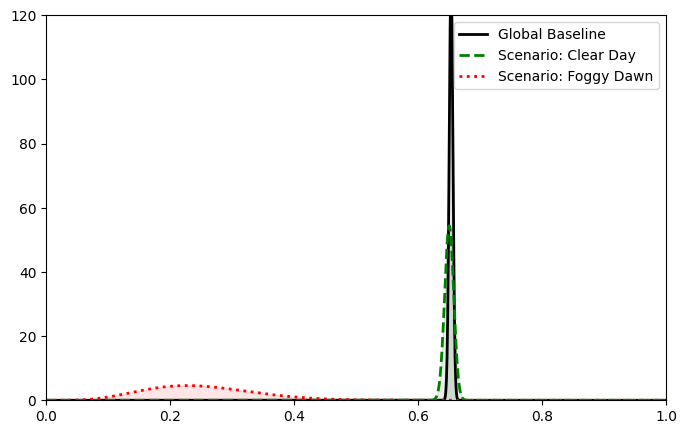}
    \caption{Beta PDFs for the \texttt{person} class. The global pooled estimate's narrow peak (black) conceals the performance gap between a benign scenario (green) and an adverse one (red).}
    \label{fig:rq2_combined}
\end{figure}

Figure~\ref{fig:rq2_combined} visualizes this contrast. The global pooled estimate and \emph{Clear\,+\,Daytime} distributions overlap as narrow peaks near $p \approx 0.65$, while the \emph{Foggy\,+\,Dawn/dusk} distribution is wide and shifted leftward -- both lower in expectation and less certain. A single pooled recall figure is therefore neither representative of benign conditions (where the system performs adequately) nor of adverse ones (where it does not).

\begin{figure}[t]
    \centering
    \includegraphics[width=1\columnwidth]{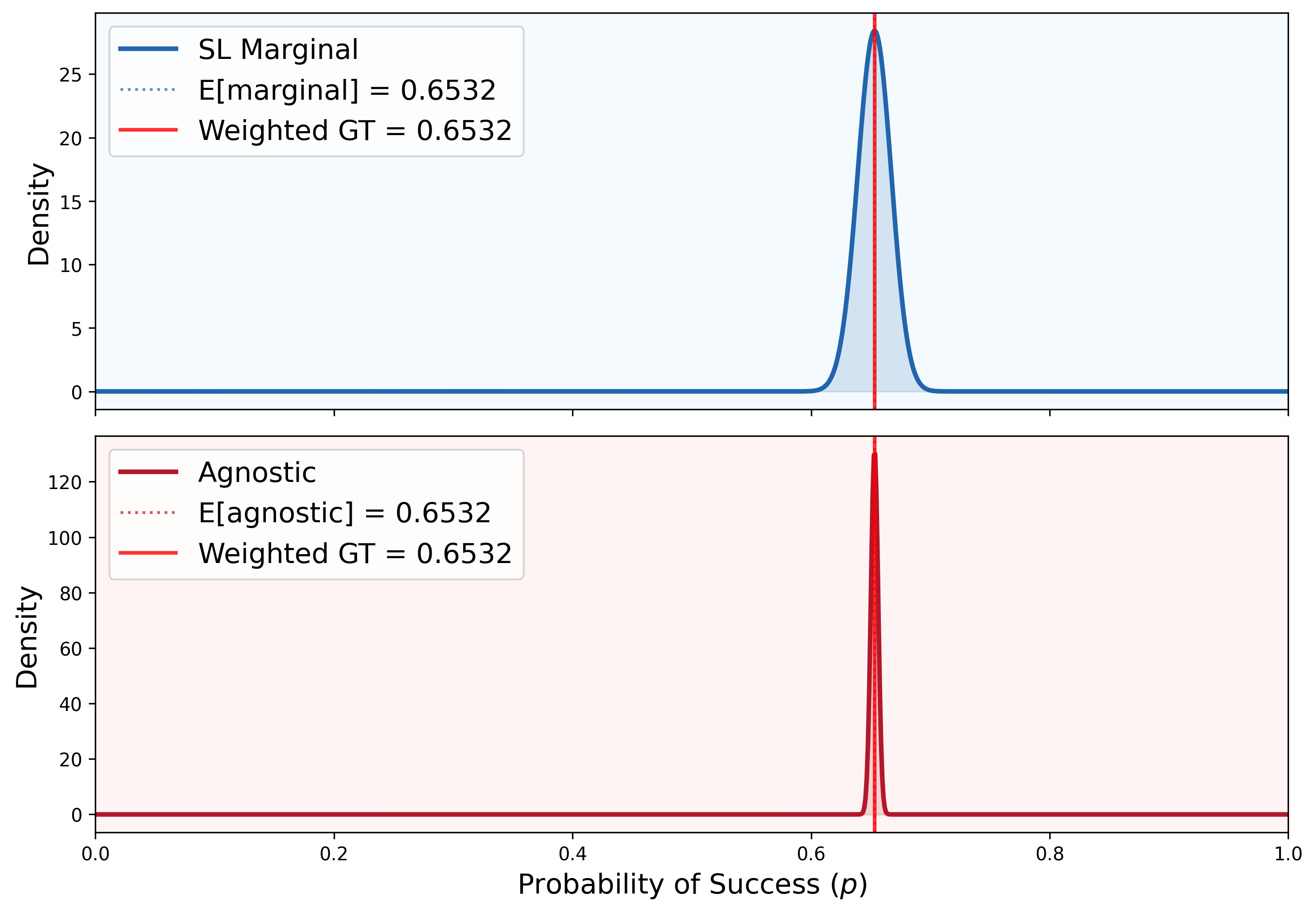}
    \caption{Marginal (top, blue) vs.\ context-agnostic pooled guarantee (bottom, red). The marginal distribution is wider and reflects the compositional uncertainty that pooling conceals.}
    \label{fig:rq2_marginal}
\end{figure}

\paragraph{Marginal vs.\ global pooled guarantees}
When the situational opinions are aggregated via SL deduction into a single marginal guarantee $\omega_{G\|S}$ and compared with the global pooled opinion (Figure~\ref{fig:rq2_marginal}, Table~\ref{tab:marginal_vs_agnostic}), the expected guarantee $E(Safe)$ is identical ($0.6532$ for the person class). The critical difference is the uncertainty mass: the marginal carries $u = 1.74 \times 10^{-3}$, roughly $21\times$ the global pooled value of $u = 8.10 \times 10^{-5}$. This gap arises because the pooled view compresses two additional sources of uncertainty into a single narrow posterior: (i)~\emph{conditional uncertainty} from data-scarce situations, whose wide per-situation posteriors are averaged away by the pooled count, and (ii)~\emph{context uncertainty} from the finite-sample estimate of how often each situation occurs in operation. The global pooled opinion still reflects finite-sample uncertainty via its $u$, but only at this aggregated level and without distinguishing where the evidence comes from. This behaviour is consistent across classes (with results for Bike, Car, and Truck shown in Table~\ref{tab:marginal_vs_agnostic}). 

\begin{table}[h]
    \centering
    \caption{Marginal vs.\ global pooled guarantee across classes. Both share the same expected value; the SL marginal exposes ${\sim}15\text{--}25\times$ more uncertainty depending on the class.}
    \label{tab:marginal_vs_agnostic}
    \begin{tabular}{@{}l c c c c@{}}
        \toprule
        & & \multicolumn{2}{c}{\textbf{Uncertainty Mass ($u$)}} & \\
        \cmidrule(lr){3-4}
        \textbf{Class} & $\mathbf{E[x]}$ & \textbf{Marginal} & \textbf{Pooled} & \textbf{Uncertainty Ratio} \\
        \midrule
        Person & 0.6532 & $1.74 \times 10^{-3}$ & $8.10 \times 10^{-5}$ & \textbf{21.4$\times$} \\
        Bike   & 0.5695 & $1.52 \times 10^{-2}$ & $1.00 \times 10^{-3}$ & \textbf{15.2$\times$} \\
        Car    & 0.7775 & $2.40 \times 10^{-4}$ & $1.00 \times 10^{-5}$ &\textbf{24.6$\times$} \\
        Truck  & 0.6575 & $4.82 \times 10^{-3}$ & $2.30 \times 10^{-4}$ & \textbf{21.0$\times$}\\
        \bottomrule
    \end{tabular}
\end{table}

Depending on the situation at hand, the global pooled guarantee may therefore be \emph{overconfident}: its narrow credible interval implicitly assumes constant operating conditions. The SL marginal provides a calibrated account of what is known (and what remains uncertain) about system-level reliability across a heterogeneous ODD. CUBICS thus enables more targeted decisions than a global model. An engineer can (i) identify situations where the expected recall is below a required threshold and uncertainty is low, indicating systematic weakness and the need for model improvement, and (ii) identify situations where recall appears adequate but uncertainty remains high, indicating that additional situation-targeted testing is required before deployment. In our example, the \emph{Foggy\,+\,Dawn/dusk} situation falls into the latter category, which would justify either restricting deployment in such conditions or prioritising data collection and testing there. From a system safety perspective, these per-situation and marginal guarantees can be used to derive concrete ODD restrictions, prioritise additional testing and data collection in high-uncertainty situations, and allocate the detector’s quantitative contribution to system-level risk in the safety case.

\begin{RQSummary}{}
\textbf{RQ2 Summary.} CUBICS identifies localized performance insufficiencies (e.g.\ \emph{Foggy\,+\,Dawn/dusk}: $E{=}0.25$, $u{=}0.083$) that the global pooled guarantee masks behind a narrow, overconfident estimate. The SL marginal guarantee retains substantially more uncertainty than the pooled context-agnostic view while recovering the same expected value, replacing false precision with transparent, actionable safety insights.
\end{RQSummary}

\subsection{\textbf{RQ3 (Sensitivity)}: How sensitive are the guarantees to prior choices, misclassification of context, and data scarcity?}
We assess the sensitivity of the CUBICS framework to the choice of base-rate prior, misclassification of context, and data scarcity as follows: 
\paragraph{Prior Choice} To investigate the effect of the choice of prior on the results, we vary the prior and observe the effect on the situation-specific guarantees. Table \ref{tab:rq3_prior_sensitivity} demonstrates that in a data-rich ODD (\emph{$s_\text{CD}$ = Clear Day}, $N=4{,}242$), the prior has very little impact on the safety guarantee. Because  uncertainty $u$ is so low ($0.0005$), the evidence ($r, s$) overwhelms the prior and anchors the expected probability $E(Safe \mid s_\text{cd})$ at $0.65$. In contrast, the data-scarce ODD (\emph{$s_\text{fd}$ = Foggy Dawn}, $N=22$) exhibits significant 
uncertainty ($u = 0.0833$). Here, the prior becomes a stronger lever: shifting from a pessimistic ($a=0.1$) to an optimistic ($a=0.9$) prior moves the guarantee by roughly $3\%$. This behavior is safety-desirable; it ensures that in situations where empirical data is lacking, the assessment is forced to rely on explicit prior assumptions rather than making overconfident claims from insufficient evidence.

\begin{table}[htbp]
    \centering
    \caption{RQ3: Sensitivity to the Base Rate Prior ($a$) in Data-Rich vs. Data-Scarce ODDs}
    \label{tab:rq3_prior_sensitivity}
    \resizebox{\columnwidth}{!}{%
    \begin{tabular}{@{}l c c@{}}
        \toprule
        \textbf{Prior ($a$)} &
        \begin{tabular}{@{}c@{}}
            \textbf{Clear Day ($N=4{,}242$)} \\
            $E(Safe \mid s_\text{cd})$
        \end{tabular} &
        \begin{tabular}{@{}c@{}}
            \textbf{Foggy Dawn ($N=22$)} \\
            $E(Safe \mid s_\text{fd})$
        \end{tabular} \\
        \midrule
        $0.1$ & $0.6499 \ (u=0.0005)$ & $0.2167 \ (u=0.0833)$ \\
        $0.5$ & $0.6501 \ (u=0.0005)$ & $0.2500 \ (u=0.0833)$ \\
        $0.9$ & $0.6503 \ (u=0.0005)$ & $0.2833 \ (u=0.0833)$ \\
        \bottomrule
    \end{tabular}%
    }
\end{table}

\begin{figure}[t]
    \centering
    \includegraphics[width=\columnwidth]{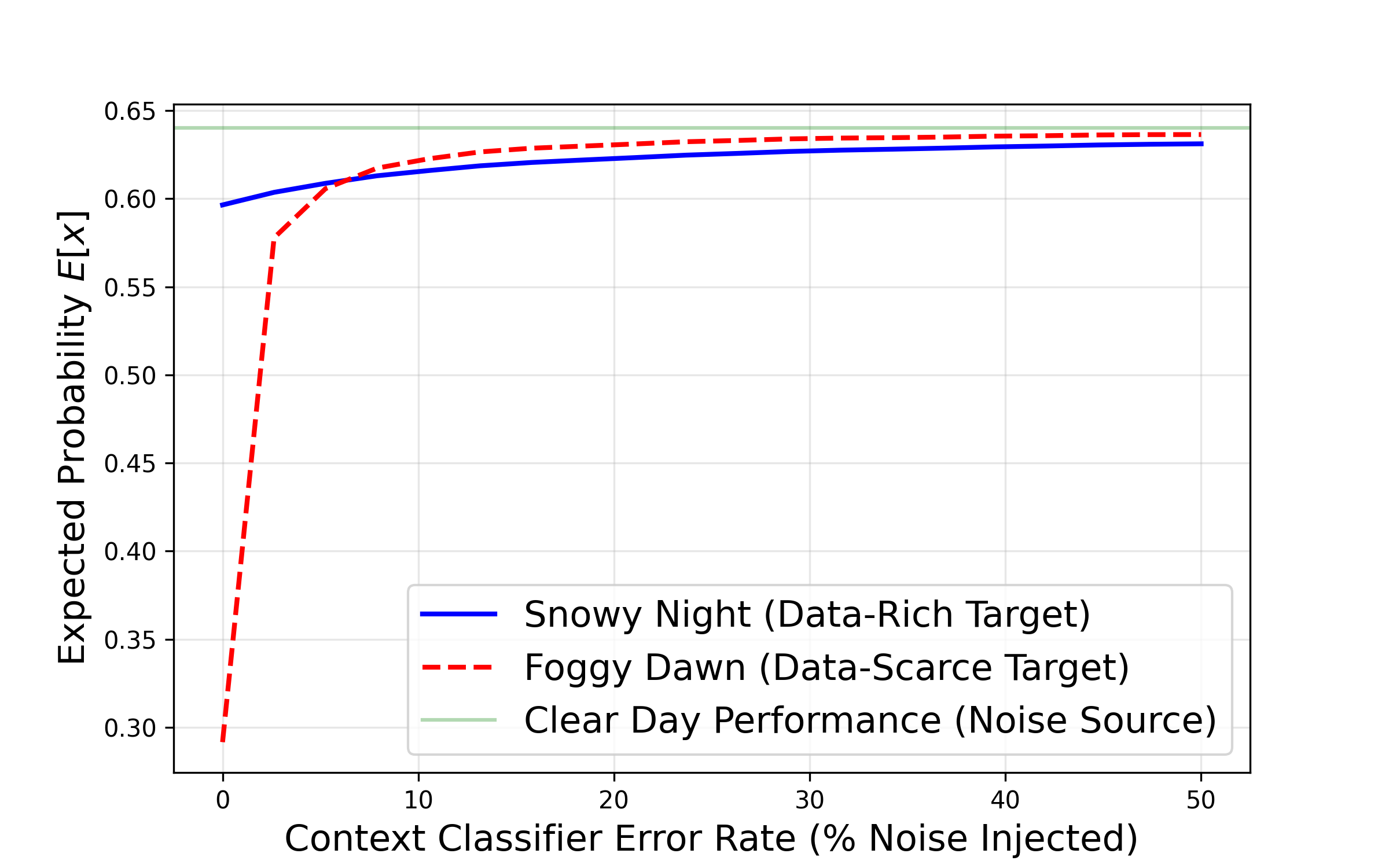}
    \caption{Sensitivity of situation-specific expected guarantee $E(Safe \mid s_{[\text{sn}/\text{fd}]})$ to context misclassification. Increasing amounts of \emph{Clear Day} evidence are incorrectly assigned to (a) a data-rich target (\emph{Snowy Night}) and (b) a data-scarce target (\emph{Foggy Dawn}), gradually inflating guarantees and reducing the distinction between situations.}
    \label{fig:rq3_context}
\end{figure}

\paragraph{Context Misclassifications} 
We simulated an imperfect context classifier by transferring varying percentages of detection evidence (from 0\% to 50\%) out of a benign, data-rich source environment (\emph{Clear Day}) and falsely attributing it to two distinct adverse Operational Design Domains (ODDs):
\begin{itemize}
    \item Data-rich target: \emph{Snowy Night} (substantial evidence).
    \item Data-scarce target: \emph{Foggy Dawn} (minimal  evidence).
\end{itemize}
By incrementally ``poisoning'' the target ODDs with evidence from the easier, higher-performing \emph{Clear Day} scenario $s_\text{cd}$, we tracked the resulting shift in the expected probability $E(Safe \mid s_{[\text{sn}/\text{fd}]})$. The results are shown in Figure \ref{fig:rq3_context}. As the proportion of misclassified evidence increases, the per-situation guarantees for the target ODDs become more optimistic and less distinct from the original source ODD. For the data-rich target, the effect on $E(Safe \mid s_\text{sn})$ remains moderate because misclassified samples are diluted by existing evidence; for the data-scarce target, even small amounts of misclassification have a visible impact on $E(Safe \mid s_\text{fd})$. In both cases, however, the associated uncertainty $u$ remains higher than in the original source ODD and increases with poisoning, and the overall behaviour gradually approaches that of a global Bernoulli model. This indicates that context misclassification mainly erodes some of the benefit of situation-specific guarantees rather than leading to overconfident, falsely precise assessments.

\paragraph{Data Scarcity} 
We describe the advantage of CUBICS in scenarios where data is scarce by examining the  edge case of detecting a truck during a foggy night. In our test set, this scenario occurred only once and resulted in a missed detection ($TP=0$, $FN=1$). A traditional frequentist evaluation yields a recall of exactly $0.0\%$ -- a brittle and overconfident penalty derived from a single sample. CUBICS avoids this deterministic trap. Recognizing the severe data scarcity ($N=1$), it assigns a dominant epistemic uncertainty mass ($u=0.6667$). Consequently, the expected guarantee is not anchored to $0\%$, but is mathematically pulled toward the non-informative base rate prior ($a=0.5$), resulting in $E(Safe \mid s_{fn})=0.3333$. This demonstrates CUBIC's capacity for \textit{graceful degradation}: rather than outputting an absolute (and statistically insignificant) safety claim, it formally bounds the localized risk with uncertainty and signals to the safety monitor that this specific ODD requires further targeted testing.

Overall, across all three experiments, realistic variations in priors, context classification quality, and sample size affect the numerical values of $E(Safe \mid s_x)$ and $u$, but did not change the qualitative risk ordering between situations.
\begin{RQSummary}{}
\textbf{RQ3 Summary}: CUBICS behaves robustly with respect to prior choices and moderate context misclassification: in data-rich situations, different reasonable priors are quickly overridden by evidence and lead to similar guarantees, whereas in data-scarce situations prior assumptions visibly influence $E(Safe \mid s_x)$ but are accompanied by high uncertainty $u$. Context misclassification gradually reduces the benefit of situation-specific guarantees and makes the model resemble a global Bernoulli view, but for realistic misclassification levels the qualitative conclusions and risk ordering between situations remain stable. Overall, CUBICS is sensitive in the intended way: it exposes uncertainty where evidence is weak, without being overly brittle to plausible modelling and sensing imperfections.
\end{RQSummary}

\section{Threats to Validity}\label{sec:threats}

\subsection{Internal}
Situational decomposition relies on correct per-image metadata; as a consequence, mislabelled metadata routes evidence to the wrong conditional opinion and distorts the corresponding situational and marginal guarantee. All opinions use the standard SL non-informative prior weight $W{=}2$; while reasonable for uninformed initialisation, it does not reflect domain-specific prior knowledge (e.g.\ from expert judgement), so posteriors for situations with sparse evidence remain sensitive to this choice. Finally, the update function treats every detection outcome within a situation as an exchangeable Bernoulli trial. In practice, FNs differ in safety relevance — missing a nearby, fully visible pedestrian is more critical than missing a distant, heavily occluded one — and failures may cluster in sub-conditions (e.g.\ dark clothing at night) that the current situation granularity does not distinguish. Incorporating severity-weighted evidence or finer situation partitions could address this, at the cost of increased data requirements.

\subsection{External Validity}
The evaluation uses TP/FN counts for a single object class, which naturally map to a binomial model. A threat to external validity is that many perception failures are continuous; extending CUBICS to such metrics would require replacing the binomial with an appropriate continuous-outcome model. Scalability is also an issue: with $k$ context dimensions of cardinality $n_i$, the number of situations grows as $\prod_i n_i$, and a realistic ODD with dozens of dimensions would cause a combinatorial explosion, with many cells having sparse evidence and guarantees dominated by the prior. Hierarchical or factored representations of the situation space are a natural mitigation. It is important to note that SL itself does not pose a computational bottleneck: the SL operators we use scale linearly with the number of situations. Furthermore, the framework assumes a fixed failure rate within each situation, although ML model performance can drift over time. Because cumulative fusion weights all historical evidence equally, outdated observations are never discounted, which may yield overconfident guarantees that no longer reflect current system behaviour. Finally, the marginal guarantee assumes that the defined situations are exhaustive. Novel or adversarial conditions not anticipated in the ODD specification fall outside the framework, and the marginal, which aggregates over defined situations weighted by observed frequencies, is blind to scenarios that have never been encountered or conceived. This limitation is shared by all scenario-based approaches but remains important: the CUBICS guarantee is only as complete as the underlying ODD partition.
\section{Related work}\label{sec:related_work}

\subsection{Continuous safety assurance and runtime monitoring}

Continuous safety assurance~\cite{schleiss2022towards} aims to keep a system’s safety argument valid throughout its lifecycle by continuously monitoring operational behaviour, detecting assurance deficits, and updating the safety case as the system or environment changes~\cite{denney2015dynamic}. Runtime monitoring and runtime assurance architectures operationalise this by observing system and environment variables, checking conformance to safety constraints, and triggering mitigation or enforcement actions when constraints might be violated~\cite{cofer2020run,hobbs2023runtime}. While such approaches can detect violations quickly and support dynamic safety cases, they typically provide qualitative signals (``constraint violated/not violated'') rather than quantified confidence in failure probabilities, and they usually lack mechanisms to (a) incorporate context distributions and (b) propagate monitoring evidence compositionally through modular arguments. Moreover, membership in scenarios or contexts (rain, night, etc.) may itself be uncertain, so updates must reflect contextual ambiguity and convert runtime evidence into quantitative, uncertainty-aware guarantees that evolve over time.

\subsection{Assurance confidence and reliability modelling}

Quantitative assurance confidence estimation aims to turn heterogeneous evidence (e.g., tests, field data, near misses) into explicit confidence levels in safety claims. In autonomous driving, naïve mileage-based demonstrations for rare catastrophic events are impractical in terms of miles required to obtain meaningful statistical bounds~\cite{kalra2016driving}, which motivates statistical and especially Bayesian approaches that explicitly model uncertainty and allow incremental updating of confidence as evidence accumulates. Standards such as ISO~26262 recognise proven-in-use style arguments but do not prescribe how to construct them for ML components whose performance is context-dependent and evolves over time. In this setting, quantitative assurance methods must cope with rare events, nonstationary failure rates, and evolving uncertainties.

Bayesian inference provides a probabilistic framework for updating hypotheses based on evidence. In safety contexts, it has been used to model component or system failure probabilities and refine them as test or field data accumulate~\cite{Guo2003,Denney2011,Hobbs2012}. For perception systems and autonomous vehicles, Bayesian reliability assessments explicitly consider that failure rates may depend on environmental conditions, moving beyond a single global Bernoulli parameter~\cite{berk2017bayesian,popov2025dynamic}. This is crucial in the rare-event regime, where context-sensitive models are needed to make efficient use of limited evidence and pure mileage-based demonstration is infeasible. Bayesian approaches also underpin assurance-oriented test planning, where sample sizes are chosen to achieve specified posterior confidence targets~\cite{wilson2021assurance}. However, prior probabilities can be subjective, evidence may be non-representative or correlated, and building and maintaining a single integrated probabilistic model for an entire system is often impractical and fragile, especially for ML-heavy systems. These limitations motivate more modular and context-structured approaches.

Subjective Logic (SL)~\cite{josang2016subjective} is an evidential reasoning framework that extends classical probability theory with explicit representation of uncertainty and base rates, and provides operators to combine and transform probabilistic opinions. It offers a convenient bridge between statistical models (e.g., Beta/Dirichlet distributions) and argumentation-level reasoning about confidence and defeaters. In assurance confidence estimation, SL has been used to reinterpret Baconian-style confidence as Beta distributions and visualise it in an opinion triangle~\cite{duan2016representation}, to make formal inferences within assurance arguments~\cite{yuan2017subjective}, to model relationships and dependencies between evidential artefacts~\cite{herd2024trust,burton2024uncertainty}, to model the behaviour of defeaters~\cite{herd2025defeaters}, and to formalise safety contracts and derive argument structures~\cite{herd2024deductive}. Building on this work, CUBICS can be seen as a concrete instantiation of the SL contract formalism for ML components: a single, context-agnostic binary contract is refined into a set of situation-indexed guarantees over context-dependent failure behaviour and equipped with an explicit Bayesian update scheme driven by field and test data. This positions SL as a suitable intermediate calculus between Bayesian statistical models and safety-case style reasoning.

In the context of continuous safety assurance, these properties are particularly valuable. SL enables assurance confidence to be updated as operational data and concerns evolve, without requiring a single monolithic system-level probabilistic model. Statistical evidence (e.g., Beta–binomial updates over pass/fail data) can be mapped into SL opinions, modified by defeaters or external information, and mapped back to probabilistic representations that feed into modular safety contracts. This interplay between Bayesian statistics and SL underpins CUBICS: situation-specific probabilistic guarantees and SL-based update rules yield quantitative, uncertainty-aware assurance confidence in a modular, context-aware manner.

Prior work already showed that reliability claims should be weighted by how systems are actually used. Musa defines reliability assessment around a quantitative characterisation of expected use, while later Bayesian approaches model reliability over input/context‑space partitions and explicitly treat uncertainty and drift in the operational profile \cite{musa2002operational,pietrantuono2020reliability}. In parallel, covariate‑dependent reliability models in classical reliability engineering use environmental factors, degradation trends, or time‑varying stresses to estimate failure risk as a function of operating conditions rather than a single global rate \cite{shyur1999general,zheng2021reliability}. At the probabilistic level, CUBICS instantiates these ideas as a context‑stratified reliability model over situation‑specific failure probabilities; the next section connects this to ODD-/scenario-based safety analysis for autonomous systems.

\subsection{Context-aware and scenario-based safety analysis}

Context-aware safety analysis treats system behaviour as explicitly dependent on operating conditions represented by an ODD specification. In automated driving, the ODD is the set of operating conditions under which a driving automation system is intended to function. Scenario-based safety evaluation builds on this by structuring verification and validation around scenarios derived from or constraining the ODD. The PEGASUS methodology~\cite{winner2018pegasus} promotes systematic scenario catalogues and links scenario-based testing to safety argumentation. Complementary work on logical scenarios formalises abstraction levels (functional, logical, concrete) and parameterisation for safety validation within the ODD~\cite{weber2019framework}.

In the SOTIF context, scenario-based approaches are used to identify triggering conditions for functional insufficiencies and to construct accelerated tests that expose SOTIF-related hazards. For autonomous systems, scenario‑ and ODD‑based safety research argues that pure mileage is infeasible, and that safety evidence should be organised around scenario coverage, environmental conditions, and risk exposure frequencies \cite{lee2020identifying,weissensteiner2023operational,tang2024scenario}. Recent reviews~\cite{zhong2021survey} distinguish between ordinary and safety-critical scenarios, discuss bidirectional interaction modelling and critical scenario generation (including corner cases), and emphasise the need to quantify exposure frequencies and residual risks per scenario. A survey on risk assessment for autonomous driving by Lu et al.~\cite{lu2025risk} shows that many methods ultimately evaluate risk at the scenario level, but still struggle with (i) uncertain or evolving operational profiles, (ii) long-tail and unknown scenarios, and (iii) aggregating heterogeneous scenario evidence into modular, uncertainty-aware guarantees suitable for safety cases.

CUBICS directly addresses these gaps by (i) decomposing the ODD into \emph{situations} that play the role of logical scenarios, (ii) associating each situation with a continuously updated Bayesian guarantee over safety-relevant behaviour, and (iii) maintaining explicit beliefs over situation occurrence. This turns the scenario/ODD partition into a set of probabilistic ``cells'' whose contributions to component-level risk can be aggregated using contracts, while preserving uncertainty and supporting through-life updates from operational evidence.
\section{Conclusions and future work}\label{sec:conclusion}

This paper introduced CUBICS, a methodology for continuous, situation‑aware performance estimation of safety-relevant ML components. Building on earlier work on SL-based safety contracts~\cite{herd2024deductive}, we extend the focus to ML components whose behaviour depends on an explicit ODD. Assumptions are represented as SL opinions over context dimensions, guarantees as SL opinions over situation-conditional failure behaviour, and SL deduction is used to derive both per-situation guarantees and a marginal, context-weighted risk contribution for each component. In this work, SL is used to represent and update component-level statistical evidence in the opinion space.

At the probabilistic level, CUBICS instantiates a context-stratified hierarchical model over situation-specific failure probabilities. The contribution is not a novel theory, but the way this structure is embedded into situation-specific SL contracts is: (i) contracts are maintained per component and per situation rather than in a monolithic system model, (ii) runtime evidence is incorporated via a context-aware fractional update mechanism that respects uncertainty in ODD labelling, and (iii) the resulting per‑situation and marginal opinions can be used as building blocks in wider safety assurance arguments. Our synthetic case study showed that, under its modelling assumptions, CUBICS recovers known situation-specific reliability patterns (RQ1). The YOLO/BDD100K study demonstrated that it exposes localised performance deficits and data gaps that a global Bernoulli model hides, and yields a more accurate component-level assessment via its marginal guarantee (RQ2). Sensitivity analyses indicated that results behave robustly under reasonable changes in priors, context misclassification, and data volume, while correctly reflecting increased uncertainty where evidence is scarce (RQ3).

Several limitations suggest directions for future work. First, the current instantiation focuses on binomial failure modes for a single component; we plan to extend CUBICS to continuous-valued metrics and to small multi-component chains to demonstrate contract composition in practice. Second, the combinatorial growth of situations and the assumption of stationarity within each cell call for hierarchical or factored situation models and temporal discounting schemes compatible with SL. Finally, we have only partially exploited the resilience branch of the contract structure in~\cite{herd2024deductive}; integrating explicit resilience goals and runtime monitors is an important step towards end-to-end, through-life safety assurance.

\section*{Acknowledgment}

The research leading to these results is funded by the German Federal Ministry for Economic Affairs and Energy within the project ``Safe AI Engineering -- Sicherheitsargumentation befähigendes AI Engineering über den gesamten Lebenszyklus einer KI-Funktion''. The authors would like to thank the consortium for the successful cooperation. We also thank the anonymous reviewers for their helpful feedback.

\bibliography{CUBICS}
\bibliographystyle{abbrv}

%
\end{document}